\documentclass[letterpaper, 10 pt, conference]{ieeeconf}  

\IEEEoverridecommandlockouts                              

\usepackage{graphicx} 
\usepackage{mathptmx} 
\usepackage{times} 
\usepackage{amsmath} 
\usepackage{amssymb}  
\usepackage{hyperref}

\title{\LARGE \bf
ARTiS: Appearance-based Action Recognition in Task Space for Real-Time Human-Robot Collaboration
}

\author{Markus Eich* and Sareh Shirazi* and Gordon Wyeth$^{1}$
\thanks{* Equal contribution}
\thanks{$^{1}$The authors are with the ARC Centre of Excellence for Robotic Vision, Queensland University of Technology (QUT), Brisbane, Australia. {\tt\small \{markus.eich,s.shirazi\}@qut.edu.au}}%
\thanks{The authors acknowledge the support of the Australian Research Council through the Centre of Excellence for Robotic Vision (CE140100016)}
}

\begin{document}

\maketitle
\thispagestyle{empty}
\pagestyle{empty}

\begin{abstract}
To have a robot actively supporting a human during a collaborative task, it is crucial that robots are able to identify the current action in order to predict the next one. Common approaches make use of high-level knowledge, such as object affordances, semantics or understanding of actions in terms of pre- and post-conditions. These approaches often require hand-coded a priori knowledge, time- and resource-intensive or supervised learning techniques.

We propose to reframe this problem as an appearance-based place recognition problem. In our framework, we regard sequences of visual images of human actions as a map in analogy to the visual place recognition problem. Observing the task for the second time, our approach is able to recognize pre-observed actions in a one-shot learning approach and is thereby able to recognize the current observation in the task space. We propose two new methods for creating and aligning action observations within a task map. We compare and verify our approaches with real data of humans assembling several types of IKEA flat packs.
\end{abstract}
\section{Introduction}
 For robots and humans to be cooperative partners that can share tasks naturally and intuitively, it is essential that the robot understands the actions of the human in order to anticipate the human’s needs. Typically, approaches to human action recognition involve the collection of images of human actions that are paired with semantic labels as the basis for a supervised learning process \cite{laptev2008learning, wang2013dense, simonyan2014two}. Alternatively, a trained object detector is paired with a human pose estimator to infer or learn the task being undertaken \cite{packer2012combined,yao2011classifying}. Both approaches measure success as the ability of the system to identify the human labeled class of action being performed in the image.

For true human robot collaboration, it is desirable that the robot is able to interact with the human in a near real-time manner. The robot has to detect the ongoing action before it is finished in order to be able to help a human e.g. during an assembly task. State of the art approaches, such as methods based on dense trajectories \cite{wang2013dense}, HOG, HOF or MBH \cite{Rohrbach2012}, and also Convolutional Neural Network (CNN) specifically trained for action recognition~\cite{simonyan2014two}, provide already good action classification results on various data sets. 

The main shortcoming is that these methods can take several seconds up to several minutes to classify an action reliably. In addition, training a task-specific CNN also requires the access to a large number of videos of our desired task as well as hours of training in advance. Therefore these methods are currently not practical in real-world human robot collaborations where an instant response is needed. Regarding the training phase, it would be desirable that the robot observes a complete task only once and is able to support a human already during a second run.

\begin{figure}[tb]
    \centering{}
    \includegraphics[width=\columnwidth]{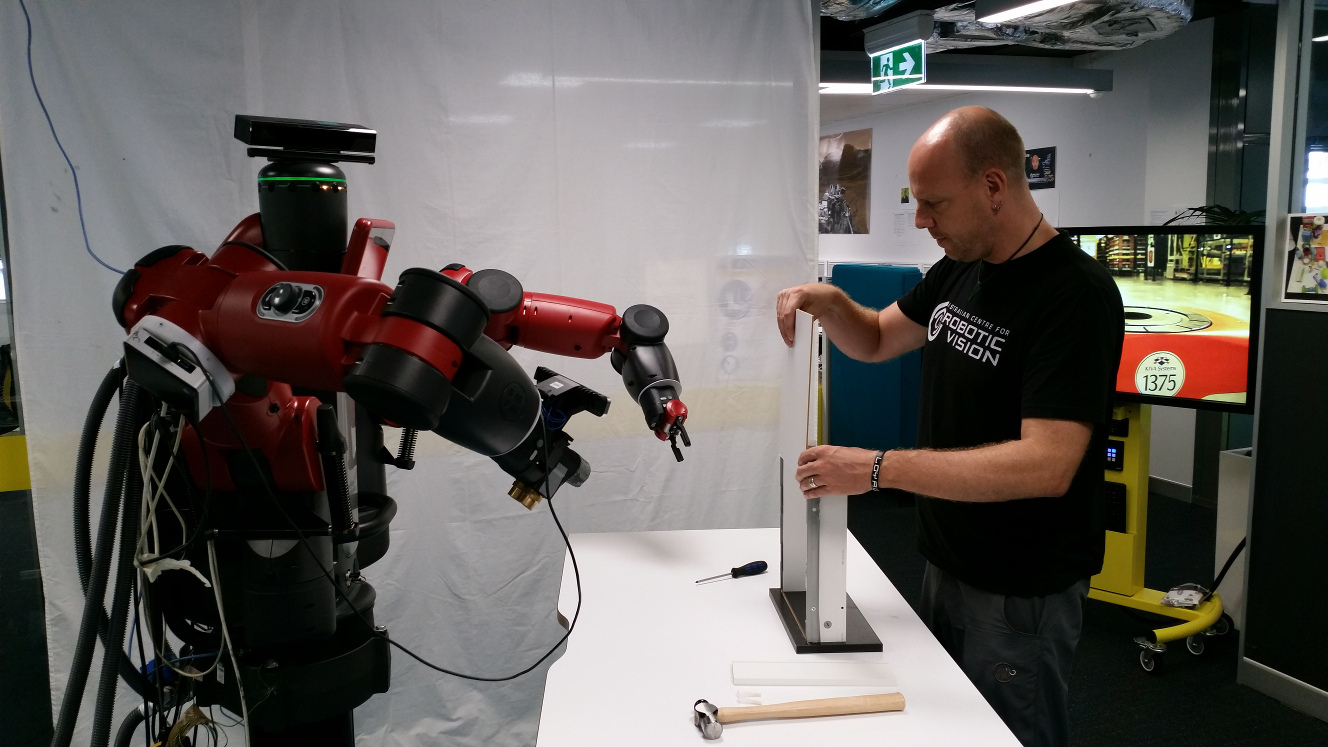}
    \caption{A typical scenario where a robot is assisting a human during an assembly task. The robot has to recognize the current action in order to be able to anticipate the next step. The robot can only be supportive if the reaction time is small. Current approaches for action recognition are not able to achieve this.}    
    \label{fig:assembly_task}
\end{figure}

In this paper, we present an on-line one-shot-learning framework. To this end, we re-frame the problem of human action recognition as identifying actions in task space. We propose an analogy between images captured during the execution of a task and images captured during movement through physical space. The analogy draws on the field of appearance-based place recognition which has shown some impressive results for localization in physical space based on one-shot learning of a route network’s appearance (\cite{Lowry2016b} for a review). The process of capturing images of the task being performed the first time creates the ``reference'' of task space. Subsequent image sequences can then be used to identify actions in the reference. This enables a robot, in contrast to state of the art methods, to interact with a human directly after the first observation. The method we are proposing in this paper is able to run on-line with a processing speed of several frames per second, similar to methods used in place recognition. We denote this system for recognising actions as ARTiS (\textbf{A}ppearance-based action \textbf{R}ecognition in \textbf{T}ask \textbf{S}pace)

We present a study of multiple people assembling different types of IKEA flat pack, ranging from drawers - see Figure~\ref{fig:assembly_task} -  to different types of tables and a TV bench. Different people follow different patterns to assemble the flat packs, both at a macro and micro level. At a macro level, an assembler might choose (for instance) to grasp the right hand or left hand side plate first. This is analogous to physically navigating down two different corridors that lead to the same place. At a micro level, the assemblers all use slightly different motions to achieve the same outcome, and have different personal appearance. This is analogous to slight variations in corridor navigation for a robot, and the variations of appearance that happen from day to day. The success of an appearance-based method for action identification in task space is measured by matching the frames for one sequence to the frames in another sequence, rather than by identifying the labeled class.

Our main contributions are: 
\begin{itemize}
    \item First, we propose a novel school of thought introducing appearance-based action recognition in task space based on one-shot learning.
    \item Second, we present two approaches enabling the robot to recognize the action (state) during the task execution process.
    \item Third, we introduce a new data set specifically designed for human-robot collaboration purposes. 
\end{itemize}
First, we intro
The paper proceeds by providing a brief review of current relevant techniques for human action recognition, and also the appearance-based methods for place recognition. The ARTiS system is proposed with two alternate methods of feature detection for generating the appearance vector and matching. The results show the performance of ARTiS with each of the feature detection methods in the context of the assembly of the several flat packs. It is important to note that our methods are unsupervised, thus being useful and scalable to different tasks.

\section{Related Work}
\label{sec:sota}
In this section, we provide a brief overview of the human action recognition techniques and appearance-based methods for place recognition.

\subsection{Human action recognition}

Fine-grained human activity recognition from unconstrained videos plays an important role for many applications such as human robot interaction. Fine-grained activities are visually similar ones that might share similar manipulations and motions for accomplishing different goals.  
There are many approaches in the literature for coarse action recognition using spatio-temporal features \cite{laptev2008learning} and dense trajectories \cite{wang2013dense}. Recent advances in using CNN have shown impressive results in image classification \cite{krizhevsky2012imagenet}. The application of CNNs has been extended to coarse action recognition \cite{simonyan2014two,yue2015beyond,rezazadegan2015}. Based on the literature, CNNs can provide a mid-level feature to capture the local motions and interactions in an activity.

Since fine-grained manipulation actions involve human object interactions, a large amount of literature focus on modelling the human and object interactions. In many works \cite{packer2012combined,prest2013explicit,yao2011classifying,wu2014human}, the authors have tried to tackle the action recognition task by modelling the contextual information between human poses, the semantics of the objects and the scenes. Using CNNs and the affordances associated with the semantics of the classified objects, in \cite{Yang2015}, a robot learns the task of cooking by recognizing the human activities at each step. 

As training the object detectors requires intensive human annotation, in \cite{zhou2015interaction} authors address this difficulty by modelling the human-object interactions without explicit object detections.

For robot assisted assembly tasks, an observation image sequence is comprised of multiple activities and multiple objects. Supervised methods would need the annotations for all the segments and objects of the training videos. Therefore, we propose an unsupervised approach.

Some methods and corresponding data sets for action recognition do already exist and are available to the community, such as the TUM data set for every activity \cite{Tenorth2009} or the popular Cooking data set \cite{Rohrbach2012}. The former makes use of a conditional random field while the latter approach makes use of HOG, HOF, and MBH to find correspondences between the sequences. The main difference between our approach and the aforementioned methods is that we use a one-shot-learning approach where the first observation of a sequence during the demonstration creates a template which is matched directly with the new observations during the exploitation phase. This makes a direct analytical comparison between the methods difficult because we have no training phase for our approach. In addition, camera view in these data sets makes it unsuitable for a collaborative robot scenario. For instance, ceiling mounted camera view in the cooking data set is not compatible with a collaborative robot's view of the working bench. That set up the motivation for collecting a new data set specifically designed for collaborative purposes.

\subsection{Appearance-based place recognition}
As we draw an analogy between appearance based localization methods in physical space and the localization in task space, we provide an overview of methods used for appearance based localization methods. Appearance-based approaches for robot navigation and place recognition are already well studied. In \cite{Cummins2008} and \cite{Nicosevici2012} bag-of-words approaches are proposed which enable a system to localize itself spatially using only sequences of 2D images. In both approaches, a visual appearance map is created in the training run. During the localization phase, the system is able to match images based on the occurrences of features and hence is able to localize within this map. 

Another approach \cite{Milford2012} proposed sequence-to-sequence matching on downscaled and patch-normalized image sequences. This allows place recognition for route navigation under changing environments. In \cite{cummins2011appearance}, in addition to descriptor-based frame similarity, they use Random sample consensus (RANSAC) to compare the feature geometry.

All the aforementioned methods use no semantics or spatial information. The identification of matches is purely based on visual similarities and the comparison of recent visual observations. In general, appearance-based approaches look for occurrences or co-occurrences of features on a frame by frame bases or within a sequence of frames. We propose in this paper to transfer the notion of appearance-based place recognition into the domain of human action recognition in task space, setting up the framework for ARTiS.

\section{ARTiS}
\label{sec:ARTiS}
Drawing on the analogy to appearance-based place recognition for robot navigation, we are interested in recognizing a recently observed action within a sequence of pre-observed actions. Similar to a mobile robot being able to recognize previously visited places based on their appearances, ARTiS enables a robot to identify the observed action in task space.

Our motivation behind ARTiS is to have a robotic system that can be easily deployed on any assistive robot. In other words, if a robot is able to identify what a human is currently doing within a sequence of a pre-observed actions in task space, it will be able to anticipate the next action. This enables a robot to actively assist a human during a task, such as an assembly task, by passing the next part or tool. 

\begin{figure*}[tb]
\centering{}
\includegraphics[width=.95\textwidth]{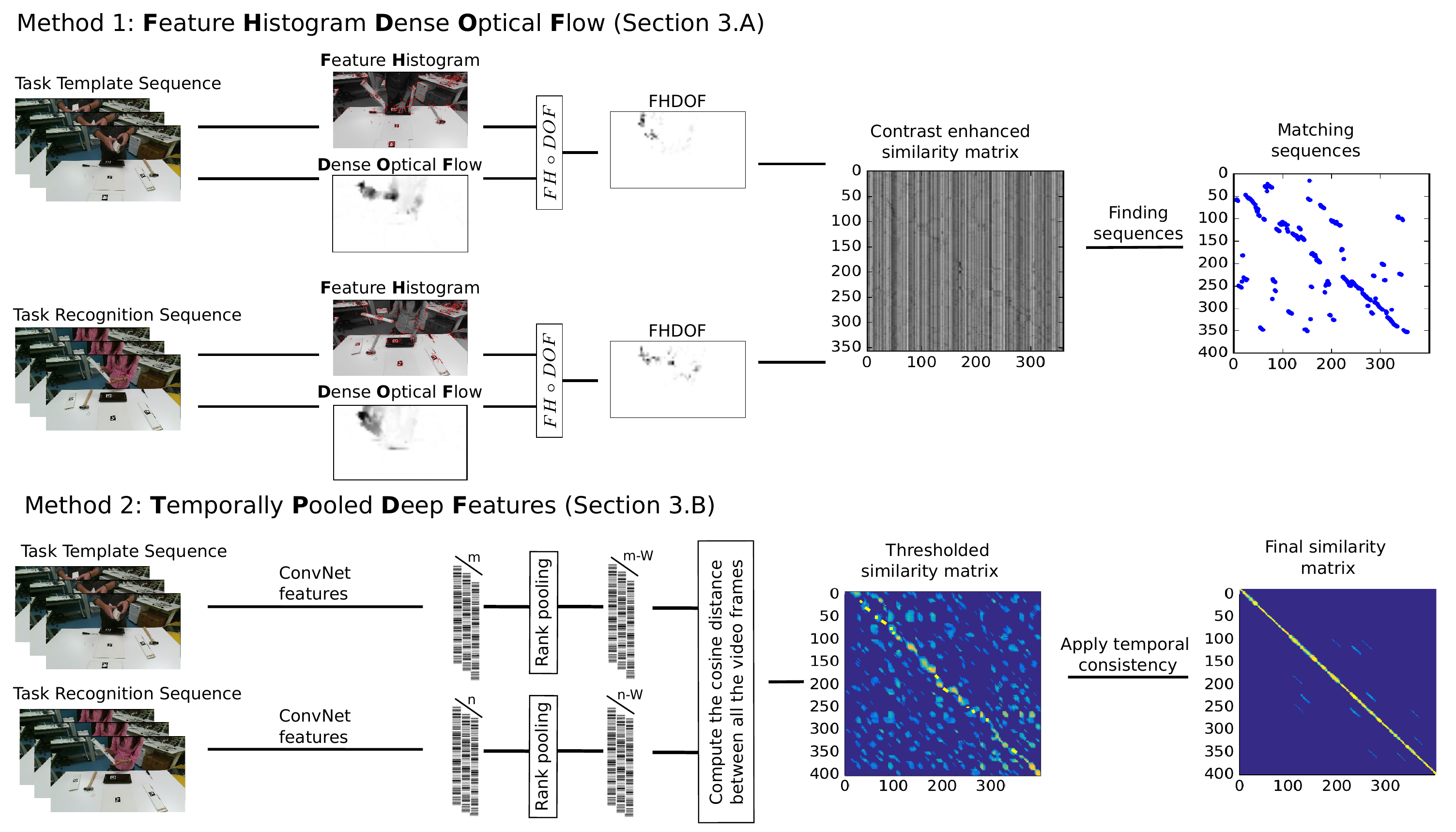}
\caption{Overview of the proposed FHDOF and TPDF methods. Each approach takes two image sequences as input. The output of each approach is a sequence of matches between the template (reference) sequence and a new observation of the same task.}
\vspace{-4mm}
\label{fig:methods_overview}
\end{figure*}

Figure~\ref{fig:assembly_task} shows an example of three different ``actions'' within the task space of an assembly task. The assembly is done by different persons with a small variation on a micro and macro level. The first observation will serve as a reference of how to assemble a workpiece.

In the second observation of the same task, which can be done by another person, we aim to identify segments of actions and sub-actions within the previously created visual template of the actions in the assembly process. Note, that there are possibly different ways to achieve the same goal. The challenge in ARTiS is that a person doing the same task might choose a different order of sub-actions at a macro level.  

One example is the assembly of the IKEA flat pack drawer used in our experiment. Although the general order of the assembly is based on the printed IKEA manual, different people follow different patterns. For instance, the drawer can be assembled using the order [\emph{attach left panel}, \emph{use screwdriver},  \emph{attach right panel}, \emph{use screwdriver}] or in the order [\emph{attach left panel}, \emph{attach right panel}, \emph{use screwdriver}]. 

Despite of having macro and micro differences, we want to assure that such variations are detected by our system. Introducing the idea of identifying actions in the task space (ARTiS), we propose two new methods. An overview of the approaches is given in Figure~\ref{fig:methods_overview}. Our two approaches are described in details below.

\subsection{Feature Histogram Dense Optical Flow (FHDOF)}
The key idea behind this method is the observation of how features are moving in the image plane during an action. The key assumption for this approach is that similar actions have similar change in appearances in the image plane. Based on how features are moving over time we can create a footprint of actions, comparable to appearance-based place recognition methods, but using temporal information instead of appearances in single frames. In this method we use a combination of feature histograms and dense optical flow to calculate changes between frames. 

\subsubsection{Feature histogram}
From the original image sequence, consisting of $1920 \times 1080$ pixel frames, a feature histogram is created. Our feature histogram is based on Speeded-Up Robust Features (SURF) \cite{Bay2008} but our method is not restricted to this method. Any method creating a density distribution of features can be used. 

Our approach uses $15 \times 15$ pixel patches are created and the number of SURF features within each patch are accumulated to a single value, resulting in a $64 \times 36$ pixel density distribution matrix. Note, that only the feature densities are taken into account and not the features itself. The reason behind is that we do not look for global matches (which can result in wrong feature alignment) but we are interested in how clusters of features are moving over time.
\subsubsection{Dense optical flow}
In order to estimate the magnitude of motion for each pixel we use the Polynomial Expansion method proposed by \cite{Farnebäck2003}. This method provides the magnitude matrix for each pixel in the images, as well as the direction of the motion which we are currently omitting. The magnitude of motion matrix is rescaled to $64\times36$ using bilinear interpolation in order to match the size of the reduced feature histogram matrix. 

The resulting FHDOF  matrix is created from the feature histogram and the magnitude of the dense optical flow field using the Schur product $FHDOF = FH \circ DOF$. This is repeated for every frame of each sequence, for the template $FHDOF_{temp}$ sequence as well as for the sequence $FHDOF_{obs}$ which is the recent observation to be recognized. 

$FHDOF$ describes the characteristic how features are moving in the image plane. Irrelevant features which are not moving between frames are removed if their magnitude of motion is zero.

\subsubsection{Task recognition using sequence based matching}
A normalized similarity $m \times n$ matrix $S$ is created from $FHDOF_{temp}$ and $FHDOF_{temp}$ where $m$ and $n$ define the number of frames in each sequence. Each value of the matrix $S(i,j)$ consists of the sum of squared pixel-wise differences between $FHDOF_{temp}$ and $FHDOF_{obs}$ created in the previous step.

We apply a simple contrast enhancement over the similarity score as proposed in \cite{Milford2012} but we achieved better results by using squared patches instead of a 1 dimensional window along the templates. The contrast enhanced similarity matrix is then defined as $S_{enh}(i,j)=\frac{S(i,j)-\bar{S}_{window}}{\sigma_{window}}$, where $\bar{S}_{window}$ defines the mean of the window around $S(i,j)$ and $\sigma_{window}$ the standart deviation within that window.

We also use the matching approach proposed in \cite{Milford2012} but instead of using reduced, patch normalized gray scale images, we take our FHDOF approach as input. The detailed description of the aforementioned method is omitted here for the sake of briefness. The reader is referred to \cite{Milford2012} for a detailed description. The overall approach is summarized in Figure~\ref{fig:methods_overview}.

\subsection{Temporally Pooled Deep Features (TPDF)}
\subsubsection{Frame level representation}
In this approach, in order to address the viewpoint dependency as one of existent problems in the field of place recognition, we propose to use a richer mid-level image representation, hence we use a layer of a pre-trained convolutional neural network to extract the features frame by frame. The network is trained on 1.28 million training images from the ImageNet 2012 classification data set, and evaluated on the 50k validation images.  Particularly, we explore the last pooling layer features of residual net with 152 layers (pool5) \cite{he2015deep} which is a 2048 dimensional vector.

\subsubsection{Sequence level representation}
Suppose we have two video sequences $\{V_a, V_b\}$, where $V_a = \left< \mathbf{r_1^a, r_2^a, \cdots r_n^a} \right>$ and $V_b = \left< \mathbf{r_1^b, r_2^b, \cdots r_m^b} \right>$. Firstly we choose a window size $W$ and then based on a recent work in action recognition \cite{fernando2016}, we pool the features across each video segment starting from $t=1$ to $t=n,m$ using the temporal rank pooling. Rank pooling is an unsupervised method that models the evolution of appearance and motion information in a video using a learning to rank methodology. In other words, the method aims to learn a function that is capable of ordering the frames of a video temporally and consequently captures the evolution of the appearance within the video. 
The result of rank pooling would be a sequence of features $P_a = \left< \mathbf{p_1^a \cdots p_{n-W}^a} \right>$ and $P_b = \left< \mathbf{p_1^b \cdots p_{m-W}^b} \right>$. 

\subsubsection{Calculating the similarity matrix}
To calculate the similarity matrix ($A$) between each pair of features, we first L2-normalize the features and then compute the cosine distance between the normalized features. To find the matched features across two videos, we choose the ones with a similarity greater than threshold $T$. We set the threshold: $T = mean \left( A \right) + 0.5 * std \left( A \right)$, where $mean \left( A \right)$ represents the average of the similarity matrix and $std \left( A \right)$ is the standard deviation. Applying this threshold would result in a new similarity matrix.

\subsubsection{Finding the best match}


To reduce the number of false positives, we apply a temporal consistency technique to solve the search problem. To this end, we assume that $\mathbf{p_i^a}$ can be matched with $\mathbf{p_j^b}$ which means their similarity should be greater than the threshold $T$. Then we focus on the relative ordering of the frames. In other words, to ensure the temporal consistency, if $\mathbf{p_{i+t}^a}$ exceeds $\mathbf{p_i^a}$ in time, its candidate match $\mathbf{p_{j+t\ensuremath{'}}^b}$ should also exceed $\mathbf{p_j^b}$ in time. An overview of TPDF is shown in Figure~\ref{fig:methods_overview}.
\section{Evaluation Methods}
\label{sec:experiments}
In this section we describe the evaluation methods we use to evaluate our FHDOF and TPDF methods. We have recorded the assembly of four different IKEA flat packs, namely the drawer KALLAX, and from the IKEA LACK series the side table, the coffee table, and the TV bench. Each run is recorded with a resolution of 1920x1080 and contains between 1500 (KALLAX) and 4000 frames (LACK coffee table). For each type of furniture we recorded 10 runs. All recordings include a half-body view of the assembler except the recording of the LACK coffee table which is assembled on the floor. The viewpoint of the camera for each sequence is approximately the same. The camera is facing the work bench as seen in Figure~\ref{fig:assembly_task}. Due to size limitations, the LACK coffee table is assembled on the ground while the other pieces are assembled on a work bench. For the KALLAX drawer, we provide two versions, one showing the full upper body including the head of the assembler and one version without the head, resulting in a total of five different runs.

The data set and the labels for the evaluation we used for our experiments are available online at \href{https://goo.gl/Oq2dYq}{https://goo.gl/Oq2dYq}.

Having the similar view point and the approximate similar distance is currently a restricting assumption of our methods. We are aware of this restriction and will discuss it in Section~\ref{sec:conclusion}. For the application of having a shared human-robot work cell, we can assume that the observing system has approximately the same view point.

In our experimental setup, parts and tools are distributed randomly on the work bench. As common for unsupervised methods,
we manually label the recorded sequences and use the ground truth information only for evaluations.
It is important to note that FHDOF and TPDF are unsupervised algorithms, we do not make use of any supervised information (any prior information about human action classes present in videos). Our main interest is identifying currently observed actions, within a sequence of previously observed actions. 

Moreover, as we mentioned in the \ref{sec:sota}, human action recognition systems are mainly dependent on numerous action videos for training, while our approach is based on one-shot learning. Therefore, to have a fair comparison, we would not include human action recognition systems in our comparisons and instead, based on the analogy between our study and place recognition, we choose SeqSLAM~\cite{Milford2012} as a baseline for comparison. 

\subsection{Sequence Labeling}
\label{sec:labeling}
To validate our approach, for each item, we have defined  multiple actions on a macro level as ground truth for the evaluation. The labels correspond to the steps displayed in the IKEA manual of the drawer. The ground truth actions for different flat packs are provided in Table~\ref{tab:primitive_actions}. Note that even if the labels are the same for different flat packs, we do not claim that the actions are interchangeable because they might have different appearance. Finding correlation between different types of flat packs will be left for future work.

\begin{table*}[htb]
    \centering
    \caption{The macro actions labeled for each type of the used IKEA flat packs. The actions are following the picture based IKEA assembly instructions.}
    \begin{tabular}{|c|c|c|c|c|c|c|c|}
        \hline
        Flat pack   & action 1              &  action 2             & action 3          & action 4          & action 5          & action 6          &  action 7 \\
        \hline
        KALLAX      & attach side panels    & attach bottom panel   & attach rear panel & use screwdriver   & use hammer        & put down          &           \\
        LACK SIDE   & attach legs           & spin legs              & rotate table      & put down          &                   &                   &           \\
        LACK COFFEE & attach legs           & spin legs              & tilt table        & attach mid panel  & use screwdriver   & put down          &           \\
        LACK TV     & attach legs           & spin legs              & rotate table      & tilt table        & attach mid panel  & use screwdriver   & put down  \\                    
        \hline  
    \end{tabular}
    \label{tab:primitive_actions}
\end{table*}

In order to verify our two methods of FHDOF and TPDF, we are interested in the correct alignment of sequences, i.e. how accurate the system can align sequences belonging to the same actions. 

For our evaluation, we create ground truths for each labeled action, defined by the start and stop frame in each sequence. This creates segments assigned to each action in the sequence space. Then, we can create the ground truth for each correct class assignment as well as for each false possible assignment. Frames which are not part of any defined actions in the ground truth are omitted from the evaluation.

\begin{figure}[tb]
    \centering{}
    \includegraphics[width=.95\columnwidth,trim={30 0 40 0},clip]{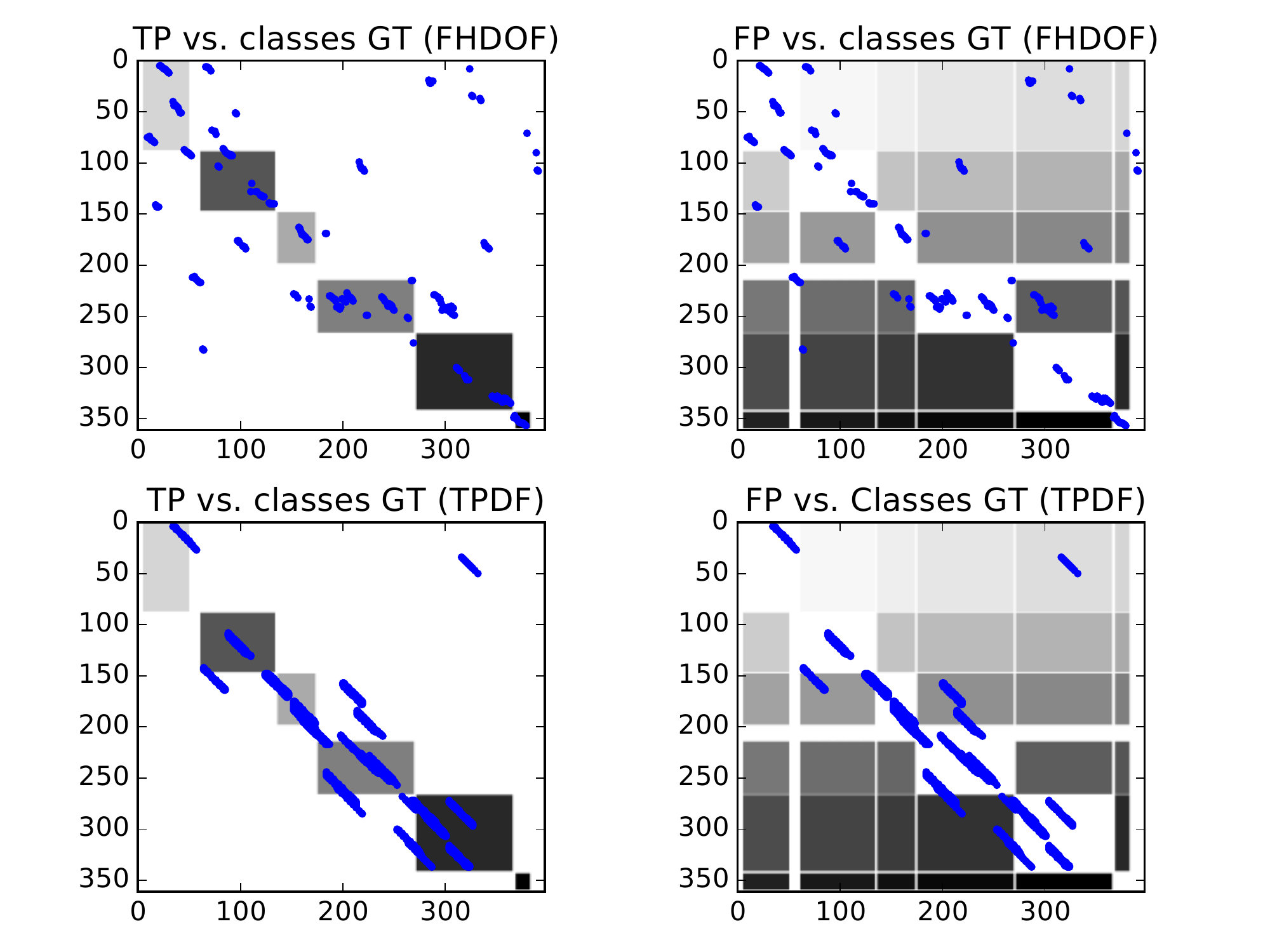}
    \caption{Matches between run 3 and run 4 of the data set compared to the labeled ground truth. Left column: Matches compared to the correct 5 action correspondences, used for the calculation of True Positives (TP). Right column: Matches compared to the 30 possible wrong action correlations used for False Positive (FP) calculation. Top row and bottom row: Direct comparison between FHDOF and TPDF. From this figure, it is directly observable that the TPDF method provides more straight lines of matches than its FHDOF counterpart which is due to preserving the temporal consistency.}
    \vspace{-4mm}
    \label{fig:ground_truth}
\end{figure}

Figure~\ref{fig:ground_truth} gives an example of how our proposed methods are evaluated. The ground truths and matches are taken from the sequences 3 and 4 of the data set. The gray shade areas in the left column represent the correspondence between frames within individual classes. Each gray area in the right column indicates between-class correspondences indicating false instances in which the predicted action class is different from the class it actually belongs to.

The matches versus the ground truth for the two sequences are provided in the figure. The top row shows the matches of our FHDOF approach, the bottom row demonstrates the results of the TPDF method.  

\begin{figure}[htb]
    \centering{}
    \includegraphics[width=.95\columnwidth,trim={0 0 0 0},clip]{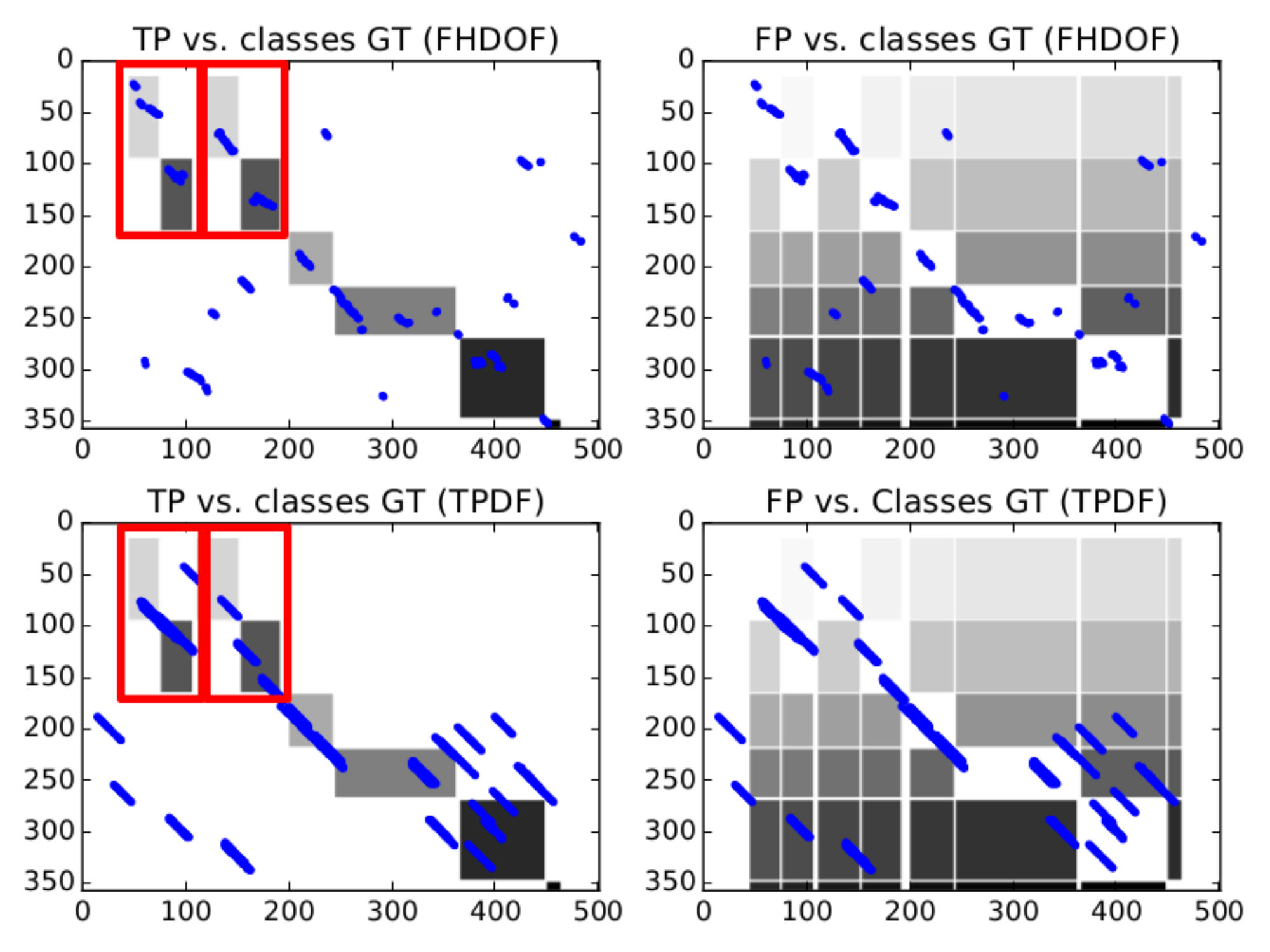}
    \caption{Matches and ground truth for two observations (run 2 and run 6 in the data set) where the order of executed actions is different, creating ''different paths`` in task space, shown as red boxes.}
    \vspace{-2mm}
    \label{fig:ground_truth_interleaved}
\end{figure}

It is important to note that matching sequences of actions in task space is not always a linear sequence to sequence match, represented as a diagonal in the similarity matrix. An example is provided in Figure~\ref{fig:ground_truth_interleaved}. In this example, the flat pack drawer was assembled using a different order of macro actions. During the task mapping reference sequence, the assembler attached the side panels and used the screwdriver to fix them to the front panel. During the recognition sequence, the assembler attached one side panel first, used the screwdriver, attached the second side panel and used the screwdriver again. This shows that different ``paths'' or different orders of macro activities can be taken to achieve the same goal. In our place recognition analogy, this would describe different traverses a robot can take within a spatial map to reach the goal.

Figure~\ref{fig:ground_truth_interleaved} suggests that our system is able to cope with such different paths in task space, using either the FHDOF approach or the TPDF approach proposed in this paper.

We follow the same evaluation method for both methods, FHDOF and TPDF. Similar to the evaluation methods in visual place recognition, we use a modified version of the frame-based evaluation in our experiments. In addition, since we aim to create a correct mapping of the actions in the assembly process and also different people have various ways of accomplishing each action, we propose to include a class-coverage-based evaluation.

The difference between class-based and frame-based evaluations is how the retrieved matches are counted as true positives (TP), false positives (FP) or false negatives (FN). Note that we do not take individual frame to frame matches into account, but only regarding sequences of matches, resulting in lines in the sequence diagram (cf. Figure~\ref{fig:ground_truth}). 

We compare the sequences of matches of each method with the ground truth areas as follows:

\subsection{Class-coverage-based evaluation}
We consider in this evaluation how precise our method is in detecting the correct class of action, provided by the labels defined as ground truth.

For each matched sequence of points, we calculate the coverage ratio of each segment of the sequences per class with respect to the correct classes (TP) or false classes (FP) (cf. Figure~\ref{fig:ground_truth} left column and right column, respectively). Each sequence of matches, whether it is located in the ``within class'' grey areas (TP) or in the ``between class'' grey areas (FP), is counted as a hit if the coverage ratio is greater than a pre-defined threshold. Note that coverage ratio indicates how much of each grey area is covered by the results and is defined as the ratio between the number of matched points in that grey area divided by the minimum length of two sequence segments correspondent to that area.

In our evaluation, we vary this specific threshold to calculate the precision/recall results of our two approaches. In our experiments, we have a maximum correct number of 6 classes. Similar to confusion matrices in statistical classification, the maximum number of false positives in our case is 30 (e.g. Action 1 (Attaching side panels of the drawer) is incorrectly aligned with Action 2 (Inserting the back panel of the drawer)).

\section{Results and Discussion}
\label{sec:results}
We test our proposed FHDOF and TPDF methods on 50 recorded image sequences of a person assembling a variety of IKEA flat packs as described above. We are primarily interested in identifying classes correctly. The approach we propose in this paper is not comparable with supervised learning and other classification methods because we do not have a training, testing and evaluation data set. 

An advantage of our proposed method compared to trajectory based methods such as \cite{wang2013dense} or methods based on HOF, HOG or MBH \cite{Rohrbach2012} is that actions can be recognized in real-time. Our proposed methods is an on-line algorithm which enables a robot to support a human during an assembly task while the task is being executed. In our experiments, we were able to achieve a processing speed of 3Hz on a standard quad-core laptop with 2.5GHz and without the use of a GPU. This allows our method to be used in real-world applications where a robot is assisting a human during the task execution without prior training on any data sets.

The way we reframe the action recognition problem in analogy to an appearance-based place recognition problem suggests that the approaches have some similarities also in the way of evaluating the results. We used the well known SeqSLAM \cite{Milford2012} method and applied it to our data set. SeqSLAM finds similarities in appearances for spatial navigation. The results shown in \ref{tab:comparison} suggest that even a more primitive approach for spatial place recognition is to a certain degree able to match actions in a series of images. The table shows that the more sophisticated approaches presented in this paper outperforms the former. We assume that this is due to the fact that SeqSLAM does only take patch-normalized images into account while our approaches considers magnitudes of motion.       

\begin{table*}[htb]
    \centering
    \caption{Comparison between our two proposed methods for each of the five different IKEA flat packs. KALLAX CU is the close-up view version which is recorded focusing only on hands, arms and parts. The other KALLAX version includes the upper body and head of the assembler We compare our two approaches with the SeqSLAM \cite{Milford2012} which is a fast online method for place recognition.}
    \begin{tabular}{|c|c|c|c|}
    \hline
    Flat pack                                       & FHDOF                                                             & TPDF            & SeqSLAM              \\    
    \hline
                &\begin{tabular}{ccc}prec.  & recall &   F1 \end{tabular}               &\begin{tabular}{ccc}prec.  & recall        & F1 \end{tabular} &\begin{tabular}{ccc}prec.  & recall        & F1 \end{tabular}
                 \\\hline
    KALLAX CU   &\begin{tabular}{ccc}0.56   &   \textbf{0.82}        &   0.66    \end{tabular}   &\begin{tabular}{ccc}\textbf{0.88}   &   0.71        &   \textbf{0.77} \end{tabular} &\begin{tabular}{ccc}0.45   &   0.63        &   0.52 \end{tabular}\\
    LACK TV     &\begin{tabular}{ccc}0.65   &   0.38        &   0.46    \end{tabular}   &\begin{tabular}{ccc}\textbf{0.86}   &   \textbf{0.77}        &   \textbf{0.79} \end{tabular} &\begin{tabular}{ccc}0.43     &   0.31        &   0.35 \end{tabular}\\        
    KALLAX      &\begin{tabular}{ccc}0.36   &   \textbf{0.57}        &   0.43    \end{tabular}   &\begin{tabular}{ccc}\textbf{0.79}   &   0.48        &   \textbf{0.55} \end{tabular} &\begin{tabular}{ccc}0.40   &   0.51        &   0.43 \end{tabular}\\
    LACK COFFEE &\begin{tabular}{ccc}0.38   &   0.34        &   0.35    \end{tabular}   &\begin{tabular}{ccc}\textbf{0.53}   &   \textbf{0.54}        &   \textbf{0.53} \end{tabular} &\begin{tabular}{ccc}0.36    &   0.17        &   0.23 \end{tabular}\\
    
    LACK SIDE   &\begin{tabular}{ccc}0.22   &   \textbf{0.24}        &   0.22    \end{tabular}   &\begin{tabular}{ccc}\textbf{0.48}   &   \textbf{0.24}        &   \textbf{0.30} \end{tabular} &\begin{tabular}{ccc}0.20   &   0.17        &   0.17 \end{tabular}\\
     \hline     
    \end{tabular}
    \label{tab:comparison}
\end{table*}



\begin{figure}[tb]
    \centering{}
    \includegraphics[width=.95\columnwidth,trim={5 0 5 0},clip]{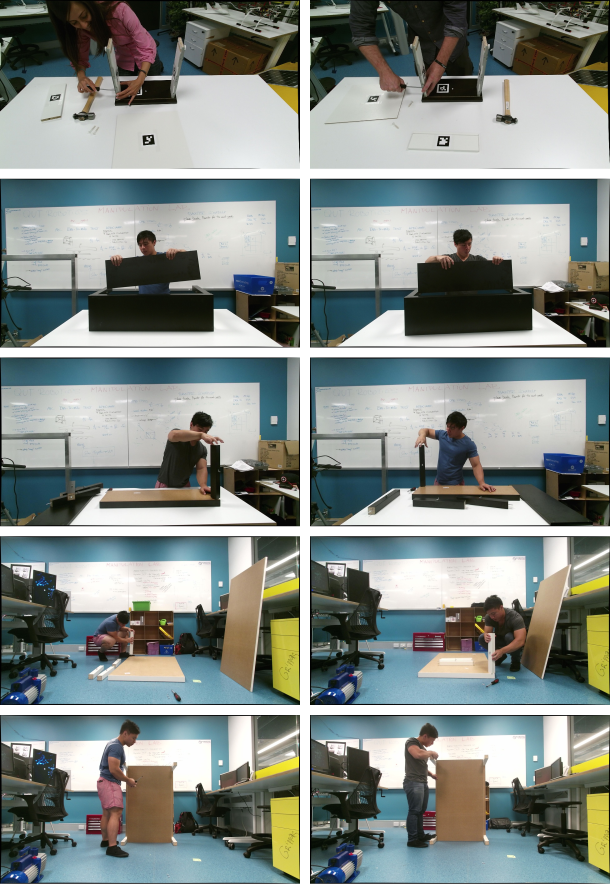}\\
    \caption{The correctly matched examples from the TPDF approach. Each row shows one correctly matched pair.}
    \vspace{-4mm}
    \label{fig:correctOnes}
\end{figure}

\begin{figure}[tb]
    \centering{}
    \includegraphics[width=.95\columnwidth,trim={5 0 5 0},clip]{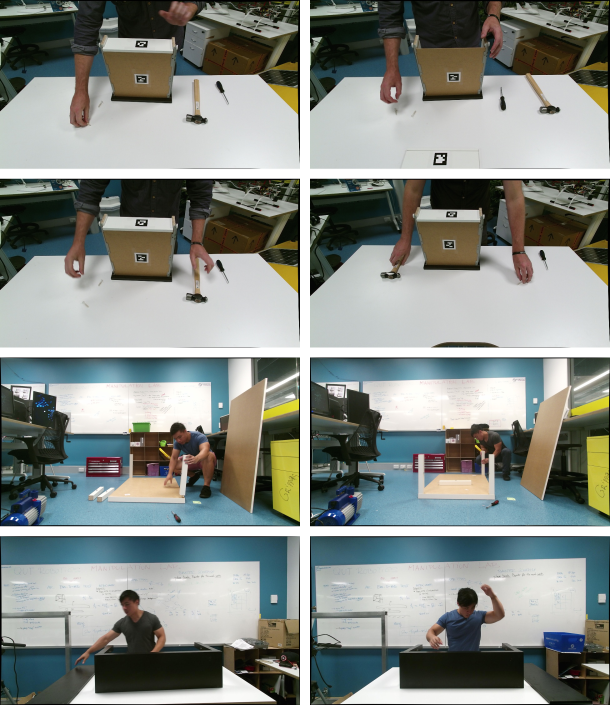}\\
    \caption{The incorrectly matched examples from the TPDF approach. Each row shows an incorrectly matched pair.}
    \vspace{-4mm}
    \label{fig:wrongOnes}
\end{figure}

Figure~\ref{fig:correctOnes} shows correctly matched examples from TPDF approach. Each row indicates different stages and actions during the assembly task for two different people.
Figure~\ref{fig:wrongOnes} also demonstrates three incorrectly matched examples from TPDF approach. Although the state of these images in the task space are different (with/without the white rear panel in the second and third row), they are both visually similar. Each frame contains key elements in almost identical positions, for instance, the assembler's hand reaching for the bolts, reaching for the hammer and drawer position.

\begin{figure}[tb]
    \centering{}
    \includegraphics[width=.49\columnwidth]{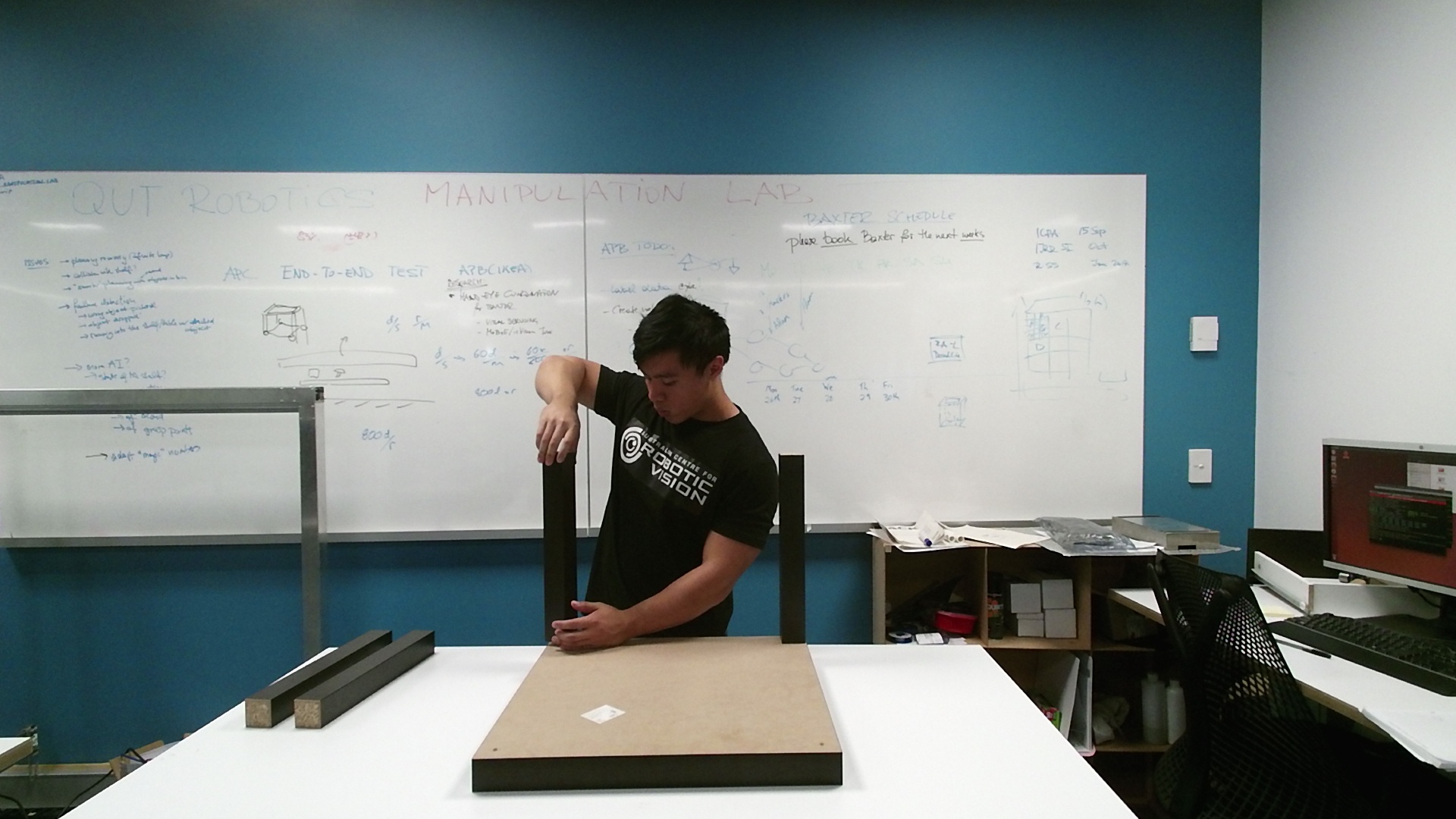}    
    \includegraphics[width=.49\columnwidth]{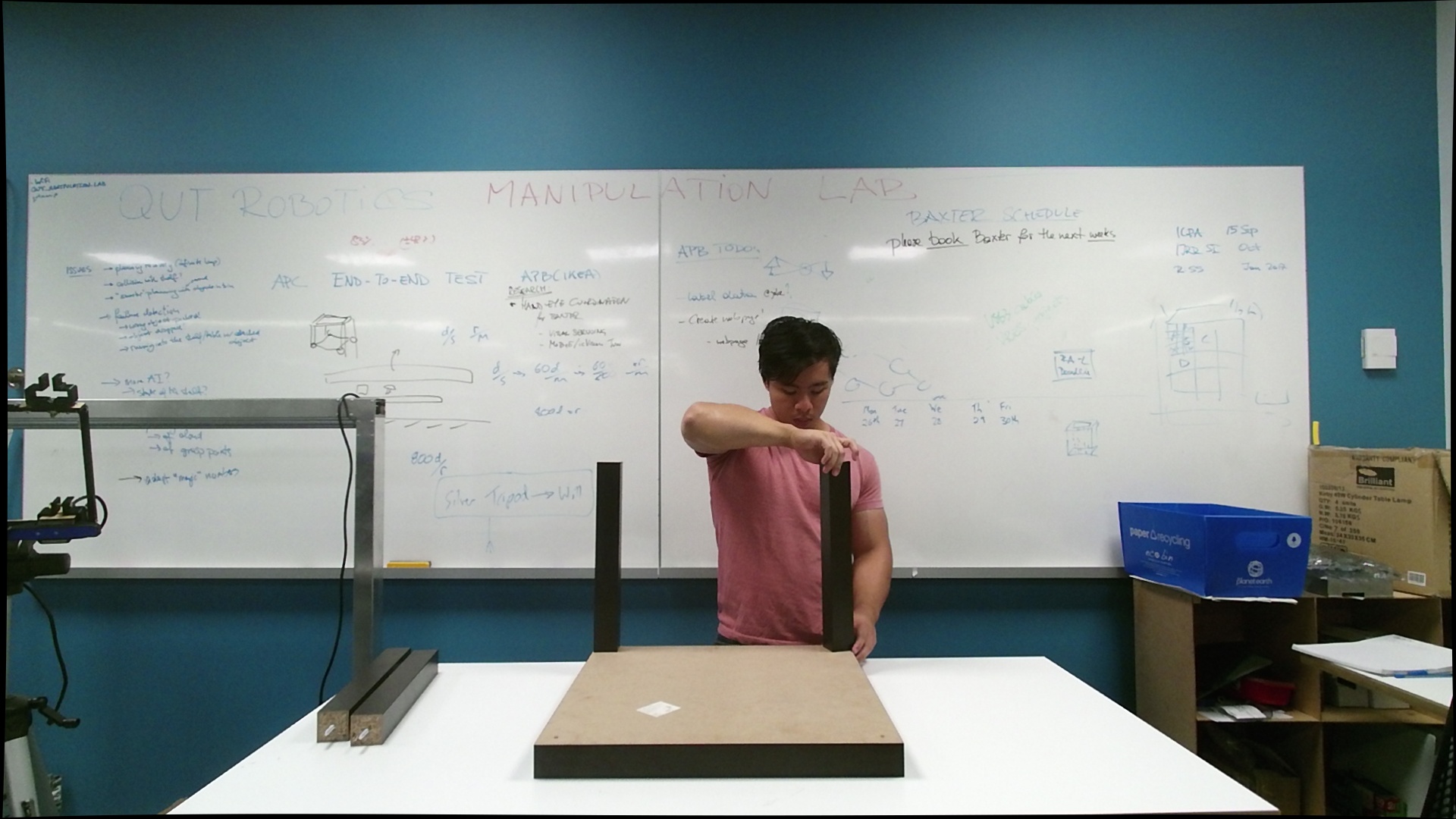}\\
    \caption{Undetected correlation using the FHDOF method because of the symmetry (mirroring) of the action \emph{Spin Leg} during the assembly of the LACK side table.}
    \vspace{-4mm}
    \label{fig:lack_leg}
\end{figure}

In our comparison, the TPDF provided slightly better results than the FHDOF method. This is mainly due to having a richer mid-level image representation as well as exploiting temporal information in feature extraction and also during the matching stage. Owing to the properties of CNNs, TPDF can partially handle variations in viewpoint, however it faces challenges in scale variations~\cite{van2016learning}.
The strength of the FHDOF is that no pre-trained model to extract the features is needed, resulting in better computability and less memory usage, making the FHDOF particularly interesting for robotic and embedded systems. 

One short-coming of our methods is the view point dependency of the system. This is due to the fact that the core of each proposed method is based on finding temporal change of 2D features, while activities take place in 3D. Viewpoint/scale dependency currently restricts the use in a scenario with fixed camera setup or a stationary robot. If the same activity takes place in different locations, it cannot be matched. This becomes obvious in a scenario with symmetries, such as assembling four times the same leg to a table in different locations (cf. LACK SIDE in Figure~\ref{fig:lack_leg} and Table~\ref{tab:comparison}. We assume that one possible solutions to deal with this issue is to look for matches in the mirrored version of the sequence. The verification of this is left open for future work. Our approach remains applicable to the scenario of a fixed robot sharing a work cell with a human, however the case of a mobile robot moving around a work place requires an investigation. 

\section{Conclusion and Future Work}
\label{sec:conclusion}
In this paper, we have introduced a new way of thinking about action recognition. We have proposed to transfer the core idea of appearance-based place recognition into the domain of appearance-based action recognition in task space, resulting in a new system we have defined as ARTiS. We have proposed two methods to recognize previously observed tasks: The first method combines feature histograms with dense optical flow in order to create a feature map of a scene (FHDOF). Sequences of FHDOF observations were used to classify actions as similar without the explicit use of semantics.

The second proposed method is based on Temporally Pooled Deep Features (TPDF) and uses pre-trained ResNet features as frame level features, temporal rank pooling for sequence level representation, combined with temporal consistency assumption. We created and published a data set which was used to evaluate our two methods.

In our future work, we will deal with the question of how to make a robot physically collaborating in real-time with a human worker based on our ARTiS architecture. 

This would require to include the semantics of tools and objects involved during each action on a macro level. This could be provided by an image classifier. Knowing the alignment between action segments in two sequences, the robot can identify the current action in the task process. By correlating actions with semantics of involved objects, the system can be enabled to predict in real-time the needed parts in the next step. In other words, if the use of a hammer is detected in one sequence, the robot can reach for the hammer if it identifies the sequence preceding the sequence where the hammer is used.

Our vision is to have a collaborative robot actively supporting a human with an assembly task by reaching for the next workpiece or tool. The perquisite should be that the robot has observed the task once or only a few times to learn the process.

\addtolength{\textheight}{-1cm}   





\bibliographystyle{IEEEtran}
\bibliography{references}

\end{document}